# Image Anomaly Detection by aggregating deep pyramidal representations

Pankaj Mishra
University of Udine, Italy
Email: mishra.pankaj@spes.uniud.it

Claudio Piciarelli
University of Udine, Italy
Email: claudio.piciarelli@uniud.it

Gian Luca Foresti
University of Udine, Italy
Email: gianluca.foresti@uniud.it

*Abstract*—Anomaly detection consists in identifying, within a dataset, those samples that significantly differ from the majority of the data, representing the normal class. It has many practical applications, e.g. ranging from defective product detection in industrial systems to medical imaging. This paper focuses on image anomaly detection using a deep neural network with multiple pyramid levels to analyze the image features at different scales. We propose a network based on encoding-decoding scheme, using a standard convolutional autoencoders, trained on normal data only in order to build a model of normality. Anomalies can be detected by the inability of the network to reconstruct its input. Experimental results show a good accuracy on MNIST, FMNIST and the recent MVTec Anomaly Detection dataset.

## I. INTRODUCTION

Anomaly detection is an application-driven problem, where the task is to identify the novelty of samples which exhibit significantly different characteristics with respect to an predefined notion of normal class. A system which can perform such task autonomously is highly in demand and its applications range from video surveillance to defect detection, medical imaging, financial transactions etc.

Only recently the topic of image anomaly detection has been investigated in the field of deep learning. The vast majority of the proposed methods rely on some for of encoding-decoding scheme, e.g. by using autoencoders, in order to train a network to reconstruct normal data [1]. The assumption is that the network is unable to correctly reconstruct anomalous images, which can be identified by direct comparison of the original and reconstructed image. However, current methods generally do not address the problem at different scales. Moreover, the comparison is often based on trivial pixel-by-pixel comparison, which is not necessarily the best approach to evaluate image similarity [2]–[6]. Finally, many papers are evaluated on toy datasets only, e.g. MNIST, which are not explicitly studied for anomaly detection problems.

In order to address these aspects, we propose a reconstruction-based pyramidal network, which uses deep autoencoders for anomaly detection. The idea is to use several parallel autoencoders with different scaling factors in order to catch features at different scale levels. To the best of our knowledge, our work is the first advocating for such multi-level design for anomaly detection. Moreover, we adopted a more sophisticated anomaly score which performs better than vanilla MSE loss adopted in many works. Finally, we tested the proposed method on the MVTec dataset [7], which is explicitly studied for anomaly detection systems.

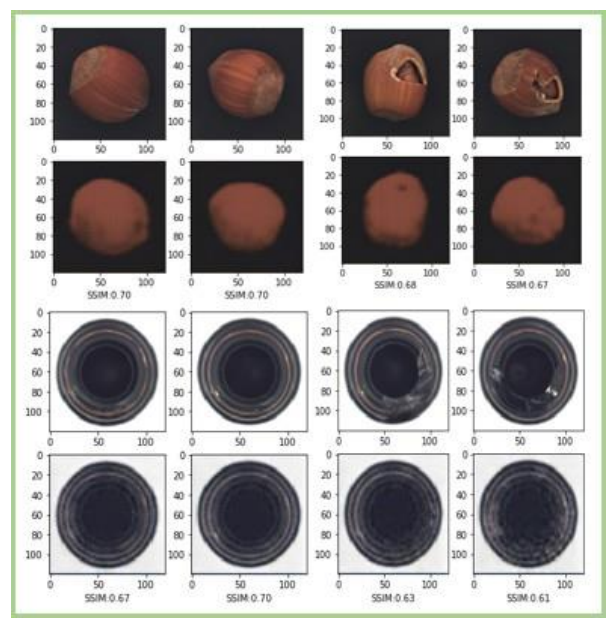

Fig. 1. Examples from the MVTec dataset. The first two columns show the original and reconstructed images for normal objects (hazelnuts and glass bottles). The last two columns show the same results for anomalous images (broken hazelnuts, defective bottles).

## II. RELATED WORK

Anomaly detection has been studied in many practical application fields, such as industrial inspection of manufactured products [8], detection of anomalous network activity in intrusion detection systems [9], medical image analysis for tumor detection [1], structural integrity check in hazardous or inaccessible environments [10], traffic analysis [11], fault-prevention in industrial sensing systems [12].

Anomaly detection can be addressed as a standard supervised binary classification problem, however in this case extreme dataset imbalance must be explicitly addressed, as in [13], because the amount of anomalous samples is typically very limited in real-world scenarios. In [14] the authors propose a transfer learning strategy to deal with class imbalance.

Most of the proposed works adopt a semi-supervised strategy: they learn a model for the normal class and try to compute a dissimilarity measure to identify the anomalies [3], [15]–[17]. This approach is semi-supervised since it requires a labeled training set consisting of normal data only, although is often improperly described in literature as unsupervised.

Some of recent proposed models exploited and relied on either parametric [18], [19] or non parametric [20], [21] approaches for density estimations of latent space for anomaly detection. Parametric models, majorly traditional machine learning techniques, uses the Gaussian density estimation techniques and Gaussian Mixture Models are recent trends with deep learning methods [22]. However, remembering an event or an instance of a particular class implies adoption of dominant features at latent space, either by a dictionary of normal prototype - as commonly the adopted methods of sparse coding approaches [23] - or by remembering the features space as in the graph based techniques [24] or the most recently used deep autoencoders [25]–[27]. All these methods tries to estimate the remembered latent features of normal class, uses either of the density estimation approaches by minimising the log likelihood loss and expects higher log likelihood loss for an anomalous sample. However, deep autoencoders have limited capacity and are not able to capture the causal factors that generates an image and are relevant to anomaly detection job. And this limits such methods.

In recent times number of works have also been done using the Generative Adversarial Network (GAN), but none of them were directly developed for anomaly detection on images. Usually, they use MSE (Mean Squared Error) or L1-norm between the pixels of ground truth and the reconstructed images, which is not how humans perceive the similarity between two images. Another problem with GAN based techniques is to find the latent vector that recovers the input image after passing through the generator [3]. Moreover, the GAN base procedures are time consuming or consist of complex multi-step training steps or are often not stable.

## III. PROPOSED MODEL

Following a global trend in deep anomaly detection, we propose a reconstruction-based approach. The basic idea is to find a low-dimension feature representation of the input image that captures its fundamental properties (the *causal factors*, as named in some works) from which the image can be reconstructed. The network thus has an encoder-decoder structure, as in standard autoencoders, which ideally models an identify function, but passing through a dimensionality-reduction bottleneck after the encoding part. The main idea is that the network, when trained on normal data, learns a mapping from input space to the low-dimensional latent space which is suitable only for normal data. If anomalous data are fed as an input to the network, their reconstruction should be poor in quality, and thus the anomalies can be detected by image comparison with the original input.

Compared to other [2], [3], [21], [28] similar works, our main contribution consists in the addition of a pyramidal level in the network structure, in order to extract features at different resolution scales. This way we increase the chances to extract features at a scale level in which the anomaly is particularly evident. Another improvement consists in the way the input and reconstructed images are compared. Most methods rely on a trivial MSE loss that, when applied to image comparison context, consists in a simple pixel-by-pixel comparison. We instead propose a high-level perceptual loss, that better models visual similarity between images.

### A. Network architecture

We propose a network to learn the manifold of the normal class by analyzing the feature representation at different scale levels using pyramidal pooling. This way the network can better find meaningful features that describe the image content at different scales, and a consequence will perform better at detecting anomalies of different sizes.

Figure 2 shows the schematic diagram of our proposed novel network. The main components of the network include:

- **Resnet18** - A pre-trained Resnet18 network (trained over imagenet dataset) is being used for deep feature extraction from images. Only the first four layers of the network have been used. The basic idea is that the network can extract generic low-level features that are meaningful for many different types of images.
- **Pyramidal Pooling Layer** - The pyramidal pooling layer thus scales the input features at different magnification levels, thus increasing the possibility that features relevant for the anomaly detection task are actually extracted. The layer takes input from the Resnet 18 block and applies an average pooling such that the output will have a unit width and height. Then it uses a convolutional layer to reduce the channel features at different scales respectively 1, 2, 3, and 6, followed by a batch normalization and ReLU activation layers. The outputs are respectively fed into four encoders.
- **Encoders** - We use four encoders, which receive their input from the pyramidal pooling layer. Each encoder is composed of a sequence of three convolutional layers which reduce the input to a feature vector in $R^8$.
- **Up-sampling Layer** - The upsampling blocks are composed of two linear layers and are used to upsample the latent features of the encoders from $R^8$ to $R^{512}$. All the upsampling blocks share the same network weights.
- **Decoder** - The decoder layer takes as input the concatenated features from upsampling layers and the output of the Resnet layer. Decoder uses 4 transposed convolutional layers to reconstruct the sample, thus giving in output an image of the same size of the input image.

Each network layer is followed by a batch normalization layer and uses the ReLU activation function except for the last layer where a sigmoid activation forces the pixel values to be in the range [0, 1]. Tables V and VI show the full structure of the encoder, upsampling and decoder layers.

### B. Objective and losses

In order to train the network we adopted a reconstruction-based approach, in which the network output is requested to be similar to the input. If the training is successful, it means that the low-dimension latent feature space in which the input is mapped after the encoders efficiently describes the visual properties of normal images. We assume that the same features will not be able to reconstruct anomalous images, which will then be identified by their higher loss. The network is trained using the following two losses:

- *Reconstruction loss*: It's a MSE loss computed between the input and the reconstructed image, i.e. $\frac{1}{WH}\|X-\hat{X}\|^2$, where $X$ is the input and $\hat{X}$ is the output of the network (reconstructed image). This is a pixel-by-pixel image comparison, widely adopted in other anomaly detection works. However, because of its intrisic pixel-level independence assumption, it fails at modeling high-level visual features.
- *Perceptual loss*: Perceptual loss [29] is a more sophisticated loss trying to catch the high-level perceptual and semantic differences between images, rather than relying on a low-level, pixel-based comparison as in the standard reconstruction loss. It is a MSE loss computed between the high-level image features obtained by a pre-trained VGG11 network using its first four layers. The loss is defined as $\frac{1}{WHC}\|F(X)-F(\hat{X})\|^2$, where $F$ is the transformation function applied through the trained four layers of VGG11 network, and $W, C, H$ are the size of the resulting feature map. The trained network is only used for the calculation of loss and the weights are not updated during training. VGG11 network is used specifically as it has a very simple network structure without any maxpool layer, which makes it ideal candidate for this work.

the proposed objective function minimizes the total loss $L$:

$$L = \frac{1}{WH}\|X-\hat{X}\|^2 + \lambda \frac{1}{WHC}\|F(X)-F(\hat{X})\|^2 \quad (1)$$

where $X$ is the input image, $\hat{X}$ is the network output, and $F$ is the non-linear function computed by the first four layers of a pre-trained VGG11 network. $\lambda$ is a weighing factor between two losses, all the experiments discussed in section IV are obtained with $\lambda = 1$.

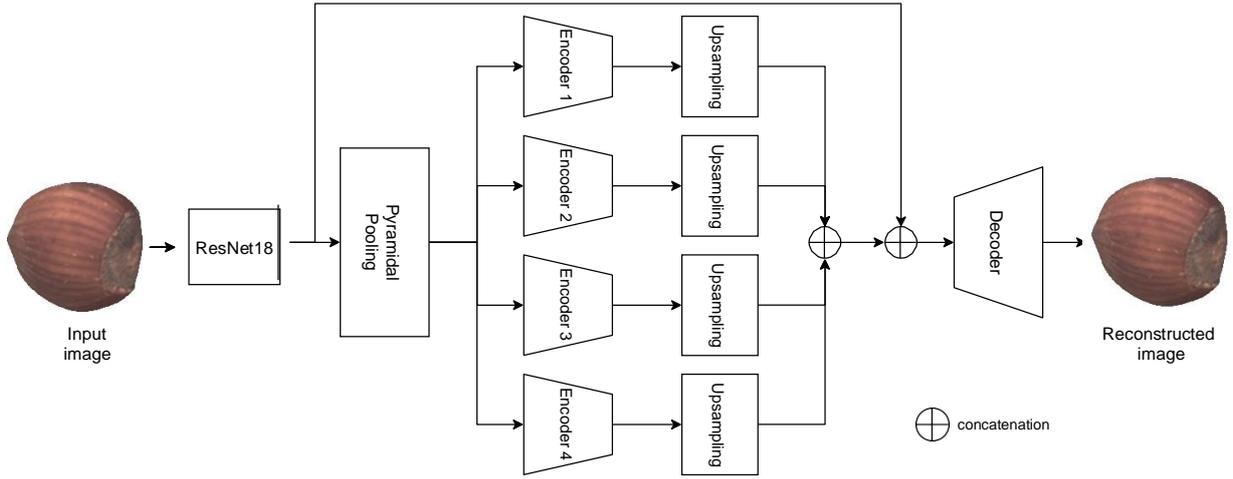

Fig. 2. Proposed network architecture. The network consists of the first levels of a Resnet18 network for feature extraction, followed by a pyramidal pooling layer that feeds 4 encoders, each one connected to an upsampling layer with shared weights, and a final decoder. The features obtained form the four upsampling layers are concatenated with the output features of resnet18.

IV. EXPERIMENTAL RESULTS

We tested the proposed model on the publicaly available standard datasets like MNIST [30] and FMNIST [31]. Although this dataset was not initially meant to be used in anomaly detection tasks, it has been widely adopted in literature to show the ability of the system to discriminate between one class, considered normal, and the other ones, considered anomalies. In addition to this we also tested our model over the recently published real-world anomaly detection dataset by MVTec [7], which contains more realistic data.

- **MNIST**: *MNIST* dataset consists of 60,000 28x28 gray images of hand written digits, grouped in 10 classes. The gray images were converted to RGB images and then passed through the network. For training, one class has been considered as the normal class while the remaining classes are considered as anomaly. Results are averaged over several runs in which each one of the original classes has been chosen as normality model.
- **Fashion MNIST**: *FMNIST* dataset composed of 60,000, 28x28 grey images of clothes from an online cloth- ing store. The imaegs were standardised similar to the MNIST before passing to the network. Table II shows our network performs better in compare to other state of the art methods like GPND [32] and OCGAN [6]. The respective ROC curve score were used for the performance measurement.
- **MVTec**: MVTec dataset contains 5354 high-resolution color images of different texture and object categories (see Fig. 1). It contains normal and anomalous images (*70 different types of anomalies*) from real world products.

Since gray-scales are quite common in industrial uses, it has 3 object categories (*zipper, screw and grid*) available solely in single channel images. As the original image sizes were large, the images were resized to *120x120* pixels before passing it to the proposed network. This size has been chosen as it maintains the structural integrity of the images such that anomalies are still visible by human eye.

Training is started by initializing the weights of the network using orthogonal initialization except the resent block, which was pretrained on imagenet, and the VGG11 block, which was pretrained on imagenet and it was kept fixed. The architectural hyper-parameters details are shown in table I.

| Adam learning Rate | 0.0001 |
|---|---|
| weight decay | 0.0001 |
| batch size | 120 |
| Epochs | 600 |

TABLE I
TRAINING HYPERPARAMETERS.

Table III shows the achieved results on the MNIST dataset. Tests have been done considering one of the classes as normal and using the remaining ones as anomalies, this has been done for each one of the 10 classes. The achieved results have been compared with standard methods such as one- class support vector machines and kernel density estimators, as well as with deep learning approaches such as denoising autoencoders, variational autoencoders [27], Pix-CNN [33] and Latent Space Autoregression [21]. The comparative results have been taken from [21]. Performance is measured using the Area Under ROC Curve (AUC) metric. As it can be seen, the proposed method achieves the best result on 6 out of 10 classes, and it has the best average result, at the same level of LSA. Table II shows the achieved results on the FMNIST dataset. Test is done similar to MNIST approach. The achieved ressults are compared with the other state-of-the art methods like GPND [32] and OCGAN [6]. Permoance is measured using AU ROC curve, averaged over 10 classes. The proposed methods performs at par and even better to the compared methods.

Table IV shows the results on the MVTec dataset over all the 16 categories, comprising both textures (carpet, grid, leather, etc.) and objects (bottle, cable, capsule, etc). Our results are compared with other deep learning anomaly detection algorithms such as autoencoders with L2 norm loss and structural similarity loss [25], the GAN-based approach AnoGAN [28], and CNN feature dictionary [34]. The comparative results have been taken from [7]. Performance is measured again using True Positive Rate (TPR) and True Negative Rate (TNR). The study follows same method with that of the compared state of the art. The proposed method achieves the best results on 9 out of 16 categories, and it reaches the best average result.

## V. ABLATION STUDY

Here we propose a set of ablation studies, in which we removed some parts of the network and the whole setup is retrained in order to see the influence of those parts on the network performance. First we try to study the effect of number of encoders. We started it with one encoder and then successively tested the cases with 2, 4, and 6 encoders. We found that the proposed model performed better in terms of anomaly detection and reconstruction capabilities with 4 and 6 autoencoders, where the performances of the two cases are substantially the same. Ablation studies have been done on the MVTec dataset ('bottle' and 'carpet', one from product and one from texture category) so that complexity of the learning domain (accuracy) and the reconstruction capacity (SSIM) can be tested. All the tests have been done with the hyperparameters kept constant (table I). The result can be seen in Figure 3 and 4.

| Network | Average AU ROC |
|---|---|
| *GPND* | 0.933 |
| *OCGAN* | 0.924 |
| *ours* | **0.936** |

TABLE II
ROC AUC FOR ANOMALY DETECTION USING FMNIST. WE REPORT THE AVERAGE VALUE FOR THE NETWORK FOR ALL THE CLASSES. RESULTS TAKEN FROM [6], [32]

| Class | OC SVM | KDE | DAE | VAE | Pix CNN GAN | LSA | Deep SVDD | Ours |
|---|---|---|---|---|---|---|---|---|
| *0* | 0.988 | 0.885 | 0.991 | 0.994 | 0.531 | 0.993 | 0.98 | **0.995** |
| *1* | **0.999** | 0.996 | **0.999** | **0.999** | 0.995 | **0.999** | 0.0.997 | **0.999** |
| *2* | 0.902 | 0.710 | 0.891 | **0.962** | 0.476 | 0.959 | 0.917 | 0.941 |
| *3* | 0.950 | 0.693 | 0.935 | 0.947 | 0.517 | **0.966** | 0.919 | **0.966** |
| *4* | 0.955 | 0.844 | 0.921 | **0.965** | 0.739 | 0.956 | 0.949 | 0.960 |
| *5* | 0.968 | 0.776 | 0.937 | 0.963 | 0.542 | 0.964 | 0.885 | **0.972** |
| *6* | 0.978 | 0.861 | 0.981 | **0.995** | 0.592 | 0.994 | 0.983 | 0.992 |
| *7* | 0.965 | 0.884 | 0.964 | 0.974 | 0.789 | 0.980 | 0.946 | **0.993** |
| *8* | 0.853 | 0.669 | 0.841 | 0.905 | 0.340 | **0.953** | 0.0.939 | 0.895 |
| *9* | 0.955 | 0.825 | 0.960 | 0.978 | 0.662 | 0.981 | 0.0.965 | **0.989** |
| *Mean* | 0.95 | 0.81 | 0.94 | 0.97 | 0.62 | **0.97** | 0.948 | **0.97** |

TABLE III
AUC RESULTS OF ANOMALY DETECTION USING MNIST. EACH ROW SHOWS THE NORMAL CLASS ON WHICH THE MODEL HAS BEEN TRAINED. COMPARATIVE RESULTS TAKEDN FROM [21]

| Class | AE SSIM | AE (L2) | Ano Gan | CNN Feat. Dict. | ours |
|---|---|---|---|---|---|
| *Carpet* | **0.43** 0.90 | 0.57 0.42 | 0.82 0.16 | 0.89 0.36 | 0.42 0.72 |
| *Grid* | 0.38 1.00 | **0.57** 0.98 | 0.90 0.12 | 0.57 0.33 | 0.86 0.53 |
| *Leather* | 0.00 0.92 | 0.06 0.82 | 0.91 0.12 | **0.63** 0.71 | 0.62 0.625 |
| *Tile* | 1.00 0.04 | **1.00** 0.54 | 0.97 0.05 | 0.97 0.44 | 0.44 0.85 |
| *Wood* | 0.84 0.82 | 1.00 0.47 | 0.89 0.47 | 0.79 0.88 | **0.85** 0.95 |
| *Bottle* | 0.85 0.90 | 0.70 0.89 | 0.95 0.43 | 1.00 0.06 | **0.84** 1.00 |
| *Cable* | 0.74 0.48 | 0.93 0.18 | 0.98 0.07 | 0.97 0.24 | **0.58** 0.89 |
| *Capsule* | 0.78 0.43 | 1.00 0.24 | 0.96 0.20 | 0.78 0.03 | **0.62** 0.74 |
| *Hazelnut* | 1.00 0.07 | 0.93 0.84 | 0.83 0.16 | 0.90 0.07 | **0.90** 0.89 |
| *Metal nut* | 1.00 0.08 | 0.68 0.77 | 0.86 0.13 | 0.55 0.74 | **0.98** 0.55 |
| *Pill* | 0.92 0.28 | 1.00 0.23 | 1.00 0.24 | 0.85 0.06 | **0.76** 0.62 |
| *Screw* | 0.95 0.06 | 0.98 0.39 | 0.41 0.28 | 0.73 0.13 | **0.73** 0.71 |
| *Toothbrush* | 0.75 0.73 | **1.00** 0.97 | 1.00 0.13 | 1.00 0.03 | 0.8 0.92 |
| *Transistor* | 1.00 0.03 | 0.97 0.45 | 0.98 0.35 | 1.00 0.15 | **0.60** 0.89 |
| *Zipper* | **1.00** 0.60 | 0.97 0.63 | 0.78 0.40 | 0.78 0.29 | 0.64 0.82 |

TABLE IV
RESULTS ON THE MVTEC DATASET. EACH ROW SHOWS THE RESULTS ACHIEVED ON A SPECIFIC CATEGORY. EACH CELL SHOWS THE BEST TNR (BOTTOM) AND TPR (TOP) VALUES. THE METHOD WITH THE HIGHEST MEAN OF THE TWO VALUES IS SHOWN IN BOLD. COMPARATIVE RESULTS TAKEN FROM LITERATURE [7].

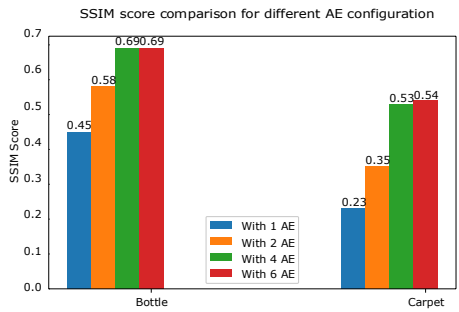

Fig. 3. SSIM comparison results for different AE configuration.

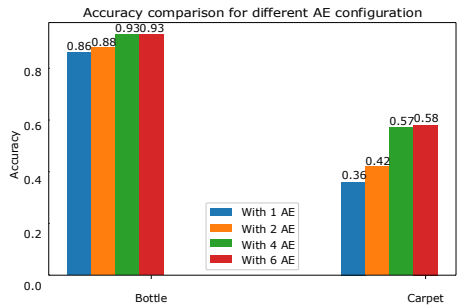

Fig. 4. Accuracy comparison results for different AE configuration.

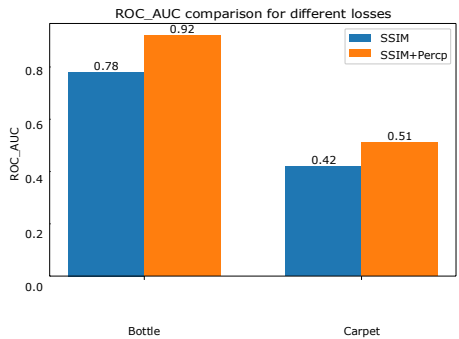

Fig. 5. ROC AUC comparison for different losses.

While average SSIM obtained from configuration having 1 and 2 encoders remained below 0.65 for bottle and 0.40 for carpet, for the normal class, the average SSIM obtained with 4 encoders remained above 0.68 for bottle and 0.53 for carpet. Also, accuracy has been tested to choose the best model configuration. The accuracy with configuration having 1 and 2 encoders remained below 85% and 36% for bottle and carpet respectively in compare to 93% and 57% obtained with 4 encoder configuration for bottle and carpet respectively. Distinctively the study showed that by adding more AE, SSIM and accuracy didn't improve much. As with the 6 autoencoder configuration the results did not improve much and remained pretty much same. Hence, we choose 4 autoencoders configuration for our further studies.

We also tried to study the effect of perceptual loss (III), over the model performance. To measure the system performance two model has been trained in following configuration: a) MSE loss only ($\lambda = 0$); b) MSE + Perceptual loss. The netowrk was trained with 4 autoencoder configuration and constant hyperparameters (see table I) for 'Bottle' and 'Carpet'. The results as measured in terms of the AUC and can be refered in Figure 5. As it can be seen, the introduction of the perceptual loss greatly enhanced the system performances.

## VI. CONCLUSIONS

In this paper we proposed a deep pyramidal network for anomaly detection. Anomalies are identified by means of a network that encodes normal images in a low-dimensional latent space and then reconstructs them, ideally modeling an identity function. Since the network is trained on normal data only, its fails at reconstructing anomalous images, which can be detected by an image similarity loss. The main contributions of this work consist in the usage of a multi-scale pyramidal approach that extract latent features at different resolutions, and the usage of a high-level perceptual loss to better compare images at feature level, rather than at pixel level. We also found that the proposed model worked best for product images (bottle, capsule, etc.), while in the case of texture images (carpet, grid, etc.) it can be further improved. Moreover, differing from many works that have been evaluated on basic datasets only such as MNIST and FMNIST, we also tested the proposed network on MVTec, a real-world dataset of defective products. Achieved results are promising and often outperform other state-of-the-art methods.

| Encoder1 | Encoder 2 | Encoder 3 | Encoder 4 | Upsampling |
|---|---|---|---|---|
| **Conv2d** **in:512,out:16,k:3,s:1,p:1** | **Conv2d** **in:256,out:16,k:3,s:1,p:1** | **Conv2d** **in:170,out:16,k:3,s:1,p:1** | **Conv2d** **in:85,out:16,k:3,s:1,p:1** | **Linear** **in:8,out:128** |
| *Batch norm* *ReLU* | *Batch norm* *ReLU* | *Batch norm* *ReLU* | *Batch norm* *ReLU* | *Batch norm* *ReLU* |
| **Conv2d** **in:16,out:8,k:3,s:1,p:1** | **Conv2d** **in:16,out:8,k:3,s:1,p:1** | **Conv2d** **in:16,out:8,k:3,s:1,p:1** | **Conv2d** **in:16,out:8,k:3,s:1,p:1** | **Linear** **in:128,out:512*$mf^2$** |
| *Batch norm* *ReLU* | *Batch norm* *ReLU* | *Batch norm* *ReLU* | *Batch norm* *ReLU* | *ReLU* |
| **Conv2d** **in:8,out:8,k:1,s:1** | **Conv2d** **in:8,out:8,k:1,s:1** | **Conv2d** **in:8,out:8,k:1,s:1** | **Conv2d** **in:8,out:8,k:1,s:1** | |

TABLE V

ENCODERS AND UP-SAMPLING LAYER ARCHITECTURE: IN, OUT, K, S, P MEANS IN CHANNEL, OUT CHANNEL, KERNEL, STRIDE AND PADDING RESPECTIVELY. 'MF' IS MULTIPLYING FACTOR WHICH IS CALCULATED AS 0.5*(OUTPUT HEIGHT OF RESNET18 FEATURES)

| MNIST, FMNIST | Mvtech |
|---|---|
| ConvTranspose2d in:64,out:16:k:5,s:1p:1 | ConvTranspose2d in:1024,out:16:k:3,s:2p:1 |
| *Batch norm* *ReLU* | *Batch norm* *ReLU* |
| ConvTranspose2d in:16,out:32:k:5,s:1 | ConvTranspose2d in:16,out:32:k:3,s:2p:1 |
| *Batch norm* *ReLU* | *Batch norm* *ReLU* |
| ConvTranspose2d in:32,out:32:k:6,s:1 | ConvTranspose2d in:32,out:32:k:4,s:2 |
| *Batch norm* *ReLU* | *Batch norm* *ReLU* |
| ConvTranspose2d in:32,out:32:k:6,s:1 | ConvTranspose2d in:32,out:3:k:4,s:2,p:1 |
| *Batch norm* *ReLU* | Tanh |
| ConvTranspose2d in:32,out:3:k:5,s:1 | |
| Tanh | |

TABLE VI

DECODER STRUCTURE